\documentclass[11pt]{article}
\usepackage[a4paper, total={6.5in, 9in}, left=1in, top=1in, right=1in, bottom=1in]{geometry}
\usepackage{xcolor}
\usepackage{amsmath,amssymb,amsfonts}
\usepackage{algorithmic}
\usepackage{graphicx}
\usepackage{hyperref}

\title{Semantic Caching of Contextual Summaries for Efficient Question-Answering with Language Models}
\author{Camille Couturier\textsuperscript{1}, Spyros Mastorakis\textsuperscript{1}, Haiying Shen\textsuperscript{1,2}, \\
        Saravan Rajmohan\textsuperscript{1}, Victor Rühle\textsuperscript{1}\\
        \textsuperscript{1}Microsoft 365 Research,
        \textsuperscript{2}University of Virginia\\
        \texttt{camille.couturier@microsoft.com}}
\date{}

\begin{document}
\maketitle

\renewcommand{\thefootnote}{}
\footnotetext{This is a preprint. This paper has been accepted at ICCCN 2025, the final version will appear in the IEEE ICCCN 2025 proceedings.}
\renewcommand{\thefootnote}{\arabic{footnote}}

\begin{abstract}
Large Language Models (LLMs) are increasingly deployed across edge and cloud platforms for real-time question-answering and retrieval-augmented generation. However, processing lengthy contexts in distributed systems incurs high computational overhead, memory usage, and network bandwidth. This paper introduces a novel semantic caching approach for storing and reusing intermediate contextual summaries, enabling efficient information reuse across similar queries in LLM-based QA workflows. Our method reduces redundant computations by up to 50-60\% while maintaining answer accuracy comparable to full document processing, as demonstrated on NaturalQuestions, TriviaQA, and a synthetic ArXiv dataset. This approach balances computational cost and response quality, critical for real-time AI assistants.
\end{abstract}


\section{Introduction}
\label{sec:introduction}

Modern Large Language Models (LLMs) have revolutionized question answering (QA), summarization, and assistant applications. Deployments spanning mobile devices to cloud datacenters face significant challenges in terms of latency, computational cost, and network bandwidth usage~\cite{xia2024hybridracahybridretrievalaugmentedcomposition}.

Many LLM-based applications involve multi-step, chained text generation with intermediate outputs. For example, in tools like Microsoft Copilot, when a user requests a summary of a conference meeting, the system processes related files such as transcripts and chat messages through a series of LLM calls before generating a response. Similarly, in QA scenarios, a retrieved document is first summarized (via one LLM call), then combined with the user's query to produce an answer (via a second LLM call). Hosting such LLM-powered systems or relying on LLM APIs can be computationally expensive and resource-intensive, as model and request volumes scale.

\begin{figure}[th]
    \centering
    \includegraphics{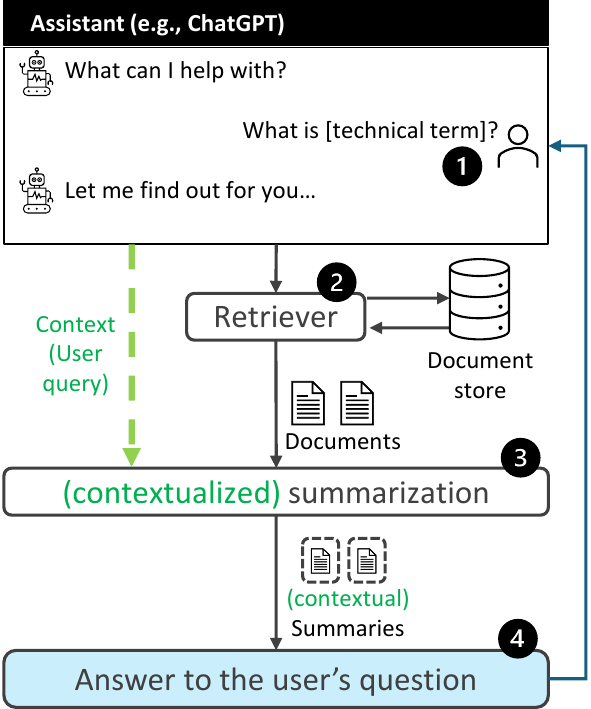}
    \caption{Simplified diagram of a chat-based assistant system with document retrieval: a user asks a question (1), the system retrieves relevant documents (2), summarizes them (3), and generates an answer (4). Summaries can be either query-agnostic (non-contextual) or query-aware (contextual). Such systems often involve multiple steps where intermediate outputs feed subsequent steps.}
    \label{fig:assistant}
\end{figure}

Numerous methods have been proposed to optimize resource usage, reduce monetary costs, and improve response time. Among these, semantic caching has emerged as a promising approach. Semantic caching stores inputs and LLM outputs as key-value pairs, retrieving and reusing previously generated outputs for future similar inputs~\cite{bang2023gptcache,codefuse-aimodelcache_2024,hf_semantic_cache}. While existing solutions focus on caching end-to-end query outputs or exact matches, little attention has been paid to intermediate data sharing among distinct requests in multi-step LLM workflows.
In a chained setup, intermediate outputs -- often discarded after use -- can be valuable for subsequent queries. For instance, summaries generated during document processing may be reusable across different user requests.
Strategies to compress intermediate data such as retrieved documents include general (non-contextual) summaries, contextual summaries tailored to specific queries, and selective extraction of key document paragraphs~\cite{Survey-2024}.

Intermediate caching offers distinct advantages over end-to-end query caching in Retrieval-Augmented Generation (RAG) systems. By storing results at intermediate stages, systems gain flexibility to handle partial document updates and varying user access while avoiding redundant computations. This approach reduces latency, enhances efficiency, and conserves computational resources—making it suitable for high-volume applications or environments with limited processing capabilities.

Semantic caching has diverse applications, such as customer support systems, where agents frequently access similar information to address common inquiries. In legal analysis, cached summaries of case law expedite responses to related questions. Personal assistant applications can leverage cached summaries of user preferences and routines for efficient, personalized interactions. Semantic caching can also support personalized learning in educational tools by reusing summaries tailored to a student's prior questions. In each case, semantic caching can enhance system performance by reducing processing overhead and response time.

This paper investigates the potential of semantic caching for intermediate contextual summaries in document-based QA tasks. Our main contributions include 1) introducing a semantic caching mechanism for reusing contextual summaries in multi-step QA workflows; 2) comparing query-agnostic and query-aware summarization strategies with other reference processing methods; and 3) evaluating the impact of semantic caching on efficiency and answer utility using publicly available QA datasets.

We use publicly available document-based QA datasets to emulate personal assistant scenarios where users ask questions and the system searches, summarizes and answers using relevant documents.
We show that semantic caching bridges the utility-cost tradeoff gap between processing the whole document or producing fresh contextual summaries for each query (high cost, high utility) and relying on general, noncontextual summaries (low cost, low utility).
For example, with a cosine similarity threshold of 0.8, semantic caching achieves cache hit ratios of 0.2-0.3 for datasets with more diverse questions, and 0.5-0.6 for datasets with more similar queries, while maintaining answer quality comparable to processing full documents.

By optimizing LLM workflows with semantic caching, we provide a scalable solution for reducing computational demand and latency, facilitating practical deployment of LLM-based systems in diverse real-world applications.

\section{Related Work}

Efficient inference of LLMs for tasks such as question-answering has led to extensive research into caching mechanisms aimed at reducing computational overhead and latency. Two primary avenues have emerged: \textbf{Prompt caching} and \textbf{Semantic caching}.

\subsection{Prompt caching (KV-based approaches)}

Prompt caching techniques build upon the key-value (KV) cache mechanism that stores attention states generated during transformer-based computations, widely adopted in high-throughput generative inference systems such as vLLM~\cite{kwonEfficientMemoryManagement2023}. While traditional KV caching focuses on reusing states within a single generation process, prompt caching extends this to reuse across different queries and LLM calls.
\textbf{Prompt Cache}~\cite{gim2023prompt} introduces a modular approach for accelerating LLM inference by reusing attention states across different prompts. It identifies frequently occurring text segments (such as system prompts, templates, or contextual documents) and precomputes their attention states, which can be efficiently retrieved when these segments reappear in future prompts. Using a Prompt Markup Language (PML) to make prompt structure explicit, it ensures positional accuracy during attention state reuse. It suffers from order-dependency, making it unsuitable for RAG systems where input order frequently varies.
\textbf{CacheBlend}~\cite{yao2024cacheblend} focuses specifically on RAG systems where multiple text chunks are incorporated into inputs to provide necessary context. It speeds up the prefill phase of long LLM inputs by pre-computing the KV cache of text chunks and reusing them regardless of whether the context appears as a prefix of another LLM input. However, its adaptability to dynamic updates of the underlying documents remains unclear, as it requires consistent input structures to maintain effectiveness.
\textbf{RAGCache}~\cite{jin2024ragcache} is optimized for RAG systems, caching and organizing intermediate KV states in a hierarchical memory and overlaps retrieval with LLM inference to reduce latency. However, its order-dependent structure limits its applicability to multi-step processes.
\textbf{EPIC}~\cite{huEPICEfficientPositionIndependent2025} addresses the position-dependency limitation by introducing position-independent context caching, enabling modular reuse of KV caches regardless of token chunk positions. Nevertheless, it still faces challenges in balancing cache size with computational efficiency.

Despite these advancements, prompt caching approaches face inherent limitations due to their focus on computational state reuse rather than semantic understanding. They excel at accelerating repetitive computations but may not fully leverage semantic relationships between queries addressing similar information needs.

\subsection{Semantic Caching}

Semantic caching aims to store and reuse previously computed outputs for semantically similar inputs, focusing on meaning rather than standard cache (exact match) or KV states. Several LLM inferencing frameworks have introduced simple semantic caching mechanisms tailored for LLMs~\cite{hf_semantic_cache,bang2023gptcache,ModelCaches}.
\textbf{GPTCache}~\cite{bang2023gptcache} stores complete LLM responses and retrieves them based on query embedding similarity. While effective for general query-response caching, it suffers from three critical limitations: (1) static similarity thresholds requiring manual tuning per domain, (2) lack of support for partial cache updates when documents change, and (3) significant memory overhead from storing complete responses.
\textbf{MeanCache}~\cite{gill2024privacy} implements a user-centric caching system that identifies similar queries and caches responses locally on user devices, leveraging federated learning for privacy preservation. Its main drawbacks include device-only storage preventing knowledge sharing between users and limitation to single-turn interactions (reducing hit rates in collaborative scenarios) and limitation to single-turn interactions without conversation history context.
\textbf{SCALM}~\cite{li2024scalm} employs hierarchical semantic clustering to identify cacheable patterns and dynamically adjust cache storage and eviction strategies to increase cache hit rates. Compared to GPTCache, SCALM shows on average a 63\% relative increase in cache hit ratio and a 77\% relative improvement in token savings. While effective for chat-based services, it struggles with niche domains due to fixed clustering granularity and lacks optimizations for document-based tasks.
\textbf{ContextCache}~\cite{mohandoss2024context} focuses on domain-specific embedding fine-tuning for legal QA applications. While achieving significant improvements in hit rates, it requires costly per-domain retraining, limiting scalability.

\subsection{Other caching approaches}

\textbf{InstCache}~\cite{zou2024instcachepredictivecachellm} introduces a predictive caching approach that differs from traditional semantic caching. Rather than storing embeddings of previous inputs or outputs, it pre-populates a cache with predicted user instructions using instruction-aligned LLMs. It leverages the observation that user instructions tend to be short, repetitive, and predictable. While achieving high hit rates, its reliance on prediction algorithms limits its adaptability in dynamic environments where user queries evolve significantly over time.
\textbf{Cache-Augmented Generation (CAG)}~\cite{chanDontRAGWhen2025} preloads all relevant knowledge into the LLM's extended context window when the knowledge base is of manageable size -- unlike conventional RAG systems that retrieve information in real-time. This approach eliminates retrieval latency, minimizes retrieval errors, and reduces system complexity. However, it is primarily effective for applications with limited knowledge needs that can fit within the model's context window, becoming less practical as the knowledge base grows.

\subsection{Our approach vs. existing methods}

Our work distinguishes itself in several key ways.
We specifically target \textbf{RAG-based applications}, focusing on document-based QA workflows.
Unlike methods that cache either complete responses (GPTCache, MeanCache, SCALM) or low-level model states (Prompt Cache, CacheBlend, RAGCache, EPIC), our approach \textbf{caches compressed, intermediate contextual summaries}, reducing storage requirements while maintaining accuracy.
Our method is \textbf{order-independent}, allowing flexibility in handling queries with varying structures -- a significant advantage over order-sensitive approaches like Prompt Cache and RAGCache which rely on specific token sequences.
Similar to other semantic caching approaches, our method leverages \textbf{query similarity} for determining cache hits, but with adaptive thresholds based on domain characteristics.
Our system uniquely offers \textbf{adaptability to dynamic updates}, enabling efficient handling of document changes without invalidating the entire cache. This is particularly valuable in real-world scenarios where documents are frequently updated or modified.
In contrast to CAG, which requires preloading all knowledge into the LLM's context window and becomes impractical with growing knowledge bases, our approach maintains scalability by selectively caching only the most relevant contextual summaries.
Table~\ref{tab:comparison} summarizes the differences and similarities among these methods, highlighting our approach's advantages in caching efficiency and flexibility.

\begin{table*}[htb!]
\centering
\caption{Comparison of design properties of our work and other prior related work.}
\label{tab:comparison}
\resizebox{\textwidth}{!}{
\begin{tabular}{|c|c|c|c|c|c|c|c|c|c|c|c|}
\hline
& \begin{tabular}[c]{@{}c@{}}Prompt Cache\\\cite{gim2023prompt}\end{tabular} & \begin{tabular}[c]{@{}c@{}}CacheBlend\\\cite{yao2024cacheblend}\end{tabular} & \begin{tabular}[c]{@{}c@{}}RAGCache\\\cite{jin2024ragcache}\end{tabular} & \begin{tabular}[c]{@{}c@{}}EPIC\\\cite{huEPICEfficientPositionIndependent2025}\end{tabular} & \begin{tabular}[c]{@{}c@{}}GPTCache\\\cite{bang2023gptcache}\end{tabular} & \begin{tabular}[c]{@{}c@{}}MeanCache\\\cite{gill2024privacy}\end{tabular} & \begin{tabular}[c]{@{}c@{}}SCALM\\\cite{li2024scalm}\end{tabular} & \begin{tabular}[c]{@{}c@{}}ContextCache\\\cite{mohandoss2024context}\end{tabular} & \begin{tabular}[c]{@{}c@{}}InstCache\\\cite{zou2024instcachepredictivecachellm}\end{tabular} & \begin{tabular}[c]{@{}c@{}}CAG\\\cite{chanDontRAGWhen2025}\end{tabular} & \textbf{Ours} \\
\hline
Caching level & {\begin{tabular}[c]{@{}c@{}}Attention\\states\end{tabular}} & {\begin{tabular}[c]{@{}c@{}}Key-value\\caches\end{tabular}} & {\begin{tabular}[c]{@{}c@{}}Key-value\\caches\end{tabular}} & {\begin{tabular}[c]{@{}c@{}}Position-\\independent\\context\end{tabular}} & {\begin{tabular}[c]{@{}c@{}}Full\\responses\end{tabular}} & {\begin{tabular}[c]{@{}c@{}}Full\\responses\end{tabular}} & {\begin{tabular}[c]{@{}c@{}}Full\\responses\end{tabular}} & {\begin{tabular}[c]{@{}c@{}}Domain-\\specific\\embeddings\end{tabular}} & {\begin{tabular}[c]{@{}c@{}}Predicted\\instructions\end{tabular}} & {\begin{tabular}[c]{@{}c@{}}Preloaded\\knowledge\\base\end{tabular}} & {\begin{tabular}[c]{@{}c@{}}\textbf{Intermediate}\\\textbf{summaries}\end{tabular}} \\
\hline
Order sensitivity & \textcolor{red}{Yes} & \textcolor{green}{No} & \textcolor{red}{Yes} & \textcolor{green}{No} & \textcolor{green}{No} & \textcolor{green}{No} & \textcolor{green}{No} & \textcolor{green}{No} & \textcolor{green}{No} & \textcolor{green}{No} & \textcolor{green}{\textbf{No}} \\
\hline
\begin{tabular}[c]{@{}c@{}}Utilizes query\\similarity\end{tabular} & \textcolor{red}{No} & \textcolor{red}{No} & \textcolor{green}{Yes} & \textcolor{red}{No} & \textcolor{green}{Yes} & \textcolor{green}{Yes} & \textcolor{green}{Yes} & \textcolor{green}{Yes} & \textcolor{green}{Yes} & \textcolor{red}{No} & \textcolor{green}{\textbf{Yes}} \\
\hline
\begin{tabular}[c]{@{}c@{}}Optimized for\\RAG systems\end{tabular} & \textcolor{red}{No} & \textcolor{green}{Yes} & \textcolor{green}{Yes} & \textcolor{green}{Yes} & \textcolor{red}{No} & \textcolor{red}{No} & \textcolor{red}{No} & \textcolor{red}{No} & \textcolor{red}{No} & \textcolor{red}{No} & \textcolor{green}{\textbf{Yes}} \\
\hline
\begin{tabular}[c]{@{}c@{}}Scalability in\\diverse scenarios\end{tabular} & \textcolor{red}{Limited} & \textcolor{green}{Yes} & \textcolor{green}{Yes} & \textcolor{green}{Yes} & \textcolor{red}{Limited} & \textcolor{red}{Limited} & \textcolor{red}{Limited} & \textcolor{red}{Limited} & \textcolor{green}{Yes} & \textcolor{red}{Limited} & \textcolor{green}{\textbf{Yes}} \\
\hline
\begin{tabular}[c]{@{}c@{}}Adaptability to\\dynamic updates\end{tabular} & \textcolor{red}{No} & \textcolor{red}{No} & \textcolor{red}{No} & \textcolor{red}{No} & \textcolor{red}{No} & \textcolor{red}{No} & \textcolor{red}{No} & \textcolor{red}{No} & \textcolor{red}{No} & \textcolor{red}{No} & \textcolor{green}{\textbf{Yes}} \\
\hline
\end{tabular}
}
\end{table*}

\section{System design and Methodology}

\subsection{Insights and System design}
\label{sec:insights}

\textbf{Insight 1. Cosine similarity captures subtle meaning in an effective way}:
Studies~\cite{reimers2019sentencebertsentenceembeddingsusing,zhelezniak2019correlationcoefficientssemantictextual} have demonstrated that cosine similarity between sentence embeddings effectively captures semantic differences between queries, often outperforming lexical metrics. In our approach, determining an appropriate similarity threshold is crucial for balancing cache efficiency and response accuracy. We employ an adaptive methodology to select this threshold, tailored to dataset and application characteristics. Specifically: (i) we decide on a acceptable performance degradation and/or on a cost reduction target; (ii) we compile a representative set of documents and query pairs and validated (``ground truth'') answers, (iii) generate embeddings (option: finetune the model encoder to the specific domain) and compute cosine similarity between the queries; and (iv) compare the performance (and/or cost) of answering with cached summaries from queries of varied similarities to conventional approaches (using the full document, or using generic, non-contextual, summaries). 
We demonstrate this trade-off in section~\ref{sec:similarity_threshold_tradeoff}.

\textbf{Insight 2. Order independence allows to focus on context rather than the order of cacheable elements}: 
Approaches based on KV caching depend on the order and position of tokens. Our selective semantic caching approach is order independent, ensuring the system can recognize and cache semantically similar content regardless of information sequence. This flexibility is crucial for handling diverse query structures and enhances caching robustness.

\textbf{Insight 3. Prioritizing user query content over static parts of the LLM input}:
For effective semantic caching, the mechanism should focus on dynamically generated user query content rather than static system prompts. System prompts typically remain unchanged across different queries, so caching them provides limited utility and can even dilute cache efficiency. By prioritizing user-specific queries, the caching mechanism can capture the most variable and relevant information, leading to higher cache hit rates and better resource utilization. This approach allows the system to deliver faster, contextually relevant responses without redundant processing of fixed system instructions.
\textbf{System design:} Our system design is based on the three insights. As shown in Figure~\ref{fig:assistant}, when the system receives a user query (step 1), it retrieves a relevant document (step 2). The system then either fetches a cached summary or generates a new one (``summarization'', step 3) and produces the answer based on this summary (step 4). Specifically, the system calculates the cosine similarity between the received question and previously received questions. If the similarity exceeds a threshold (e.g., 0.85), the system fetches the cached summary of the document and uses it to generate the answer. If no similar historical question is found, a new summary is generated, processed, and cached for future use. This approach saves time on summary generation and reduces response latency.

\subsection{Our Semantic Caching methodology}
\label{sec:methodology}

In an LLM-based assistant application, when a user submits a query, the system first searches for relevant documents and then generates a summary to answer the question. The summary can be categorized as query-agnostic (non-contextual) or query-aware (contextual). Query-agnostic summaries are generated without specific reference to the query and are based solely on the document's content, providing a general overview. In contrast, query-aware summaries are tailored to the user's query, incorporating context from both the query and the documents to provide a more precise and relevant response. See Figure~\ref{fig:assistant} for a visual demonstration of our system.

Borrowing the formulation from LLMLingua's work on prompt compression~\cite{jiang_llmlingua_2023, jiang_longllmlingua_2023, pan_llmlingua-2_2024}, we consider that a prompt $\mathbf{x}$ is composed of the system instructions $\mathbf{x}^{\text{ins}}$, the user query $\mathbf{x}^{\text{que}}$, and additional materials $\mathbf{x}^{\text{doc}}_1, \cdots, \mathbf{x}^{\text{doc}}_K$.
\textbf{With our approach, these additional materials are replaced by their summarized counterparts} $\mathbf{x}^{\text{sum}}_1, \cdots, \mathbf{x}^{\text{sum}}_K$. We leave the actual retrieval and ranking of documents aside and focus on summarizing one document at a time. Importantly, \textbf{the system instructions $\mathbf{x}^{\text{ins}}$ are not considered in the summarization process}. A prompt $\mathbf{x}$ thus simply reads as
$\mathbf{x}=(\mathbf{x}^{\text{doc}}, \mathbf{x}^{\text{que}})$.

\textbf{Query-agnostic} (\textbf{non-contextual}) summaries $\mathbf{x}^{\text{sum}_i}$ are functions of the document alone:
$\mathbf{x}^{\text{sum}_i} = f(\mathbf{x}^{\text{doc}_i})$.
\textbf{Query-aware} (\textbf{contextual}) summaries are tailored to answer the user query:
$\mathbf{x}^{\text{sum}_i} = f(\mathbf{x}^{\text{doc}_i}, \mathbf{x}^{\text{que}})$.
Also, \textbf{the cached summaries are indexed on the user queries $\mathbf{x}^{\text{que}}$, not on the full prompt $\mathbf{x}$ (selective caching)}.

\begin{figure*}[ht!]
\centering
\footnotesize
\begin{verbatim}
SystemMessage(content="""
You are an Expert at Summarizing text content and all your outputs are expertly curated. You will
ask the user for a document. You will then write a precise, detailed 200-word summary for the
document. The summary should be informative and stick to the information from the document."""),
AIMessage(content="What is the document you would like to summarize?"),
HumanMessage(content=document),
AIMessage(content="Here is a summary of the document:\n")
\end{verbatim}
\vspace{-0.1in}
\caption{Instructions used to prompt the LLM to generate a general, non-contextual summary of a document}
\label{fig:prompt_general_summary}

\vspace{0.07in}
\begin{verbatim}
SystemMessage(content="""
You are an Expert at Summarizing text content and all your outputs are expertly curated. You will
ask the user for a document, then for a question. You will then write a precise, detailed 200-word
summary for the document that will help answering the question and follow-up or related questions.
The summary you write *must* contain a precise, detailed answer to the question, *if and only if*
it is present in the document. The summary should be informative and stick to the information from
the document."""),
AIMessage(content="What is the document you would like to summarize?"),
HumanMessage(content=document),
AIMessage(content="What is the question you would like to answer?"),
HumanMessage(content=question),
AIMessage(content="Here is a summary of the document that answers your question:\n")
\end{verbatim}
\vspace{-0.1in}
\caption{Instructions used to prompt the LLM to generate a query-aware, contextual summary of a document}
\label{fig:prompt_contextual_summary}

\vspace{+0.07in}
\begin{verbatim}
SystemMessage(content="""
You are an Expert at Answering questions using text content and all your outputs are expertly
curated. You will ask the user for a document, then for a question. You will then write a concise
(max. 3 words) answer to the question, *if and only if* it is present in the document. The answer
should be informative and stick to the information from the document."""),
AIMessage(content="What is the document you would like to use?"),
HumanMessage(content=reference),
AIMessage(content="What is the question you would like to answer?"),
HumanMessage(content=question),
AIMessage(content="Here is the answer to your question:\n")
\end{verbatim}
\vspace{-0.1in}
\caption{Instructions used to prompt the LLM to generate an answer using a reference (a summary or the whole document)}
\label{fig:prompt_answer}
\end{figure*}

\section{Experimental setup}
\label{sec:experimental_setup}

To evaluate the effectiveness of semantic caching, we simulate interactions with a QA assistant over time, modeling a real-world scenario where users ask multiple questions related to a set of documents.

\subsection{Simulation design}
\label{sec:simulation_design}

The experimental design follows the QA assistant workflow mentioned in Section~\ref{sec:insights}: for each input question, a document is "retrieved"; a summary of the document is generated; and an answer is produced using the summary.
The simulations adapt this base workflow to five different answering methods:
\textbf{Full document}: The system generates an answer using the entire document each time, skipping the summarization step.
\textbf{No retrieval}: The system generates an answer using only the internal knowledge of the LLM, skipping document retrieval and summarization.
\textbf{Non-contextual summary}: The system generates a general, non-contextual summary once and reuses it for all questions.
\textbf{Contextual summary}: The system generates a fresh contextual summary tailored to the user query or reuses a cached summary of the document from a previous user query.
\textbf{Cache the answer, indexed on the entire prompt}: The system caches the answer generated for each prompt -- incl. system instructions, the document and the question -- and retrieves it when a similar prompt is received.

Figures~\ref{fig:prompt_general_summary}-\ref{fig:prompt_answer} provide the instructions used to prompt the LLM to generate a general, non-contextual summary of a document, a query-aware summary, and an answer using a reference (a summary or the whole document). 
We call the OpenAI API with GPT models~\cite{brown_language_2020,openai_GPT35turbo} through the LangChain chat module~\cite{langchain_github} to generate both summaries and answers, with a temperature of 0 and a top-p of 1. We generate summaries of lengths of (100, 200, 400) words for the non-contextual and the contextual summaries, based on the average document length in our dataset and the trade-off between compression and information preservation. 
We use the FAISS~\cite{JDH17} library to build a cache of the summaries indexed on the questions embeddings.

\subsection{Datasets}
\label{sec:datasets}

Our evaluation focuses on short-form QA (2-13 word questions) reflecting real-world usage patterns: NaturalQuestions~\cite{kwiatkowski_natural_2019} and TriviaQA~\cite{joshi_triviaqa_2017}; we also craft a synthetic dataset based on recent ArXiv preprints to test technical content handling. These datasets provide a range of question complexities and document types, ensuring robust evaluation.
To ensure realistic simulations of various queries attached to each document, we select documents with at least 10 questions (for the synthetic ArXiv dataset, we generate exactly 15 different questions per document). We also filter out documents with less than 800 words.
\begin{table}[h]
\centering
\caption{Statistics of the datasets used for our experiments. Showing (mean $\pm$ std) where applicable.}
\begin{tabular}{|l c c c|}
\hline
& \textbf{TriviaQA} & \textbf{NaturalQuestions} & \textbf{ArXiv} \\
\hline
\textbf{Documents} & & & \\
\# docs & 20 & 83 & 134 \\
\# words per doc & 3423 $\pm$ 2613 & 10824 $\pm$ 7863 & 4057 $\pm$ 930 \\
\hline
\textbf{Questions} & & & \\
Total \# questions & 398 & 1248 & 2010 \\
\# questions per doc & 20 $\pm$ 25 & 15 $\pm$ 8 & 15 $\pm$ 0 \\
\# words per question & 13 $\pm$ 5 & 9 $\pm$ 2 & 8 $\pm$ 3 \\
\hline
\textbf{Answers}: \# words & 2 $\pm$ 1 & 2 $\pm$ 2 & 3 $\pm$ 4 \\
\hline
\end{tabular}
\vspace{0.1in}
\label{tab:datasets_stats}
\end{table}
Table~\ref{tab:datasets_stats} shows the statistics of the datasets. The documents are relatively long, with an average of 11k words for NaturalQuestions. The questions are much shorter, averaging 8-13 words. The number of questions per document varies, with the ArXiv dataset having a fixed number of generated questions per document. The answers are concise, averaging 2-3 words\footnote{We chose the ``short answer'' format when possible to facilitate the evaluation of answer relevance on the relatively small available datasets.}.

\subsection{Similarity between questions}

We quantify the relatedness of questions using cosine similarity between embeddings obtained with sentence encoders -- we used \textrm{all-MiniLM-L6-v2} and \textrm{all-mpnet-base-v2} models with the SentenceTransformer library~\cite{reimers-2019-sentence-bert}.

\begin{figure}[ht]
\centering
\includegraphics[width=0.5\linewidth]{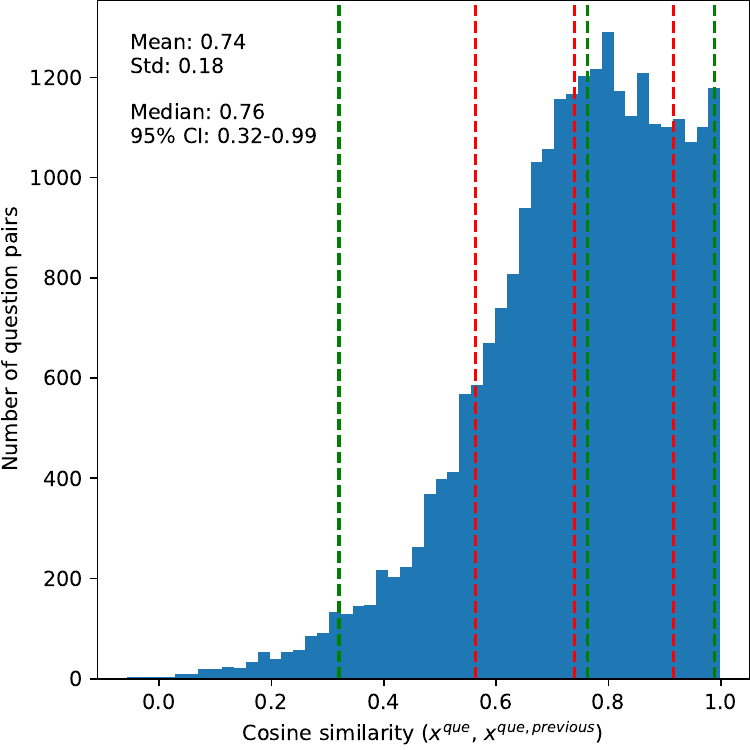}
\vspace{-0.08in}
\caption{NaturalQuestions dataset: cosine similarity between pairs of questions asked about a document.}
\label{fig:questions_similarity_NQ}
\end{figure}

Figure~\ref{fig:questions_similarity_NQ} shows the distribution of cosine similarity between per-document question pairs in NaturalQuestions. The distribution is skewed toward high similarity scores (mean 0.74, median 0.76), indicating that many questions are closely related and may share common information or context. This suggests cached summaries can be effective in reducing redundant computations. However, the wide range of similarities (standard deviation 0.18, 95\% confidence interval ranging from 0.32 to 0.99) hints that the utility of cached summaries may diminish, as they may not contain relevant information for answering significantly different questions. This underscores the importance of having a dynamic caching strategy that can adapt to varying levels of question similarity.

\subsection{Summaries}
\label{sec:triviaqa_summaries}

\begin{figure}[t]
\centering
\includegraphics[width=0.9\linewidth]{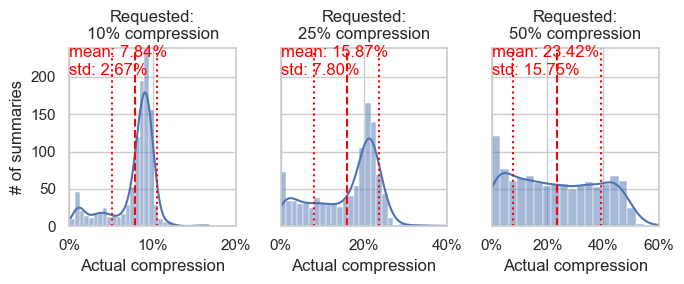}
\includegraphics[width=0.9\linewidth]{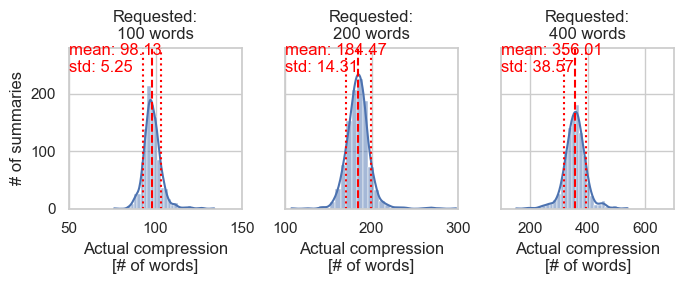}
\vspace{-0.2in}
\caption{Ratios (compared to the document length) and lengths (in terms of words) of different summaries (10\%, 25\%, 50\%, 100 words, 200 words, 400 words) obtained through GPT-4o-mini on TriviaQA documents.}
\label{fig:TQA_summaries_word_ratio}
\end{figure}

Figure~\ref{fig:TQA_summaries_word_ratio} shows the ratios and lengths of different non-contextual summaries (10\%, 25\%, 50\%, 100 words, 200 words, 400 words) obtained through GPT-4o-mini on TriviaQA documents\footnote{The distributions for contextual summaries, and for the other two datasets, are very similar.}, cf. instructions in Figures~\ref{fig:prompt_general_summary}-\ref{fig:prompt_answer}.
For highly compressed summaries (100-word, 10\% summaries), the mode (most frequent) compression is close to the requested one. However, for summaries with lower requested compression ratios (50\%, 400-word), the number of words/tokens is consistently lower than the budget, suggesting that the GPT-4o-mini-based summarization method may not be using the compression budget optimally. Additionally, when comparing ratio-based summaries to fixed-length summaries, we find that the latter follow the requested compression ratio more closely. We thus use fixed-length summaries in our experiments.

\subsection{Evaluation metrics}
\label{sec:metrics}

The effectiveness of each retrieval method is evaluated on:

\begin{itemize}
\item \textbf{Utility:} Measured as cosine similarity between generated and ground truth answers. Higher utility indicates more accurate and relevant answers.
\item \textbf{Cache hit rate:} Proportion of queries for which a cached summary is available. A higher cache hit rate indicates more effective reuse of cached data, leading to reduced computational load.
\item \textbf{Token usage:} Approximation of the cost of generating an answer, measured by the total \textbf{number of input tokens} fed to the LLM, incl. the system prompt, instructions, document, summary (if applicable) and the total \textbf{number of output tokens} generated by the LLM, \textit{i.e.}, the length of the answer and the summary (if applicable).
\item \textbf{Latency:} Time taken to produce an answer end-to-end.
\end{itemize}

\section{Results and discussion}
\label{sec:results}

\subsection{Comparative analysis of retrieval methods}
\label{sec:retrieval_methods}

\begin{figure}[t]
    \centering
    \includegraphics[width=0.99\linewidth, trim=5mm 28mm 3mm 37mm, clip]{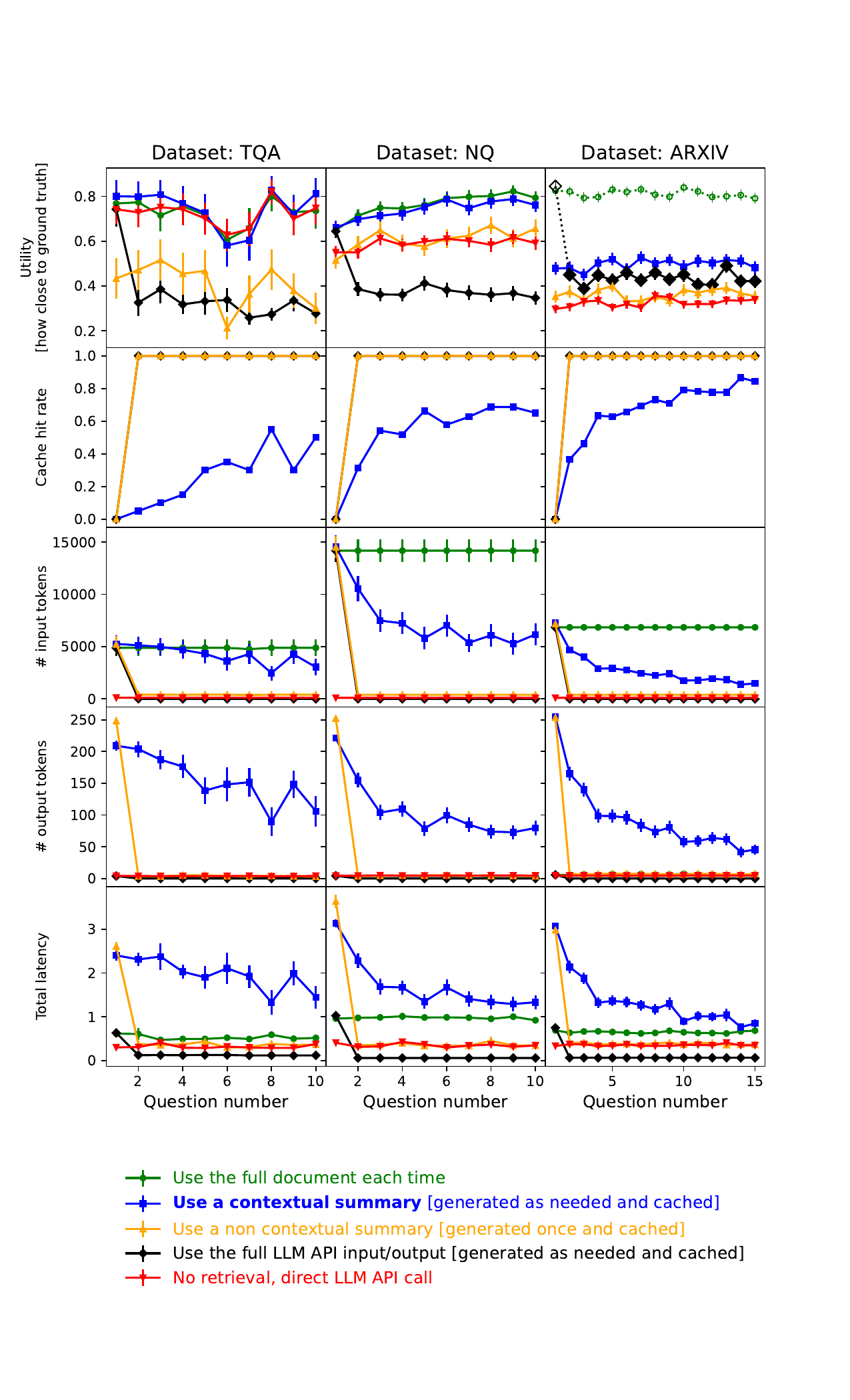}

    \vspace{-0.08in}

    \caption{Comparative analysis of retrieval methods across TriviaQA, NaturalQuestions, and synthetic ArXiv datasets, for a similarity threshold of 0.8 and summary length of 200 words. 
    Each row represents a different metric as the number of questions increases: utility, cache hit rate, input/output tokens, and latency.
    Note: For the ArXiv dataset, full document answers serve as the reference, creating artificially high utility values for this method (shown as dotted lines).}
    \label{fig:retrieval_comparison}
\end{figure}

Figure~\ref{fig:retrieval_comparison} provides a comprehensive evaluation of the five retrieval methods (colored lines) across the TriviaQA, NaturalQuestions, and synthetic ArXiv datasets.

Regarding our \textit{contextual summaries} method, results show the \textbf{utility} is very close to the ideal one (\textit{Using the full document}), indicating high accuracy in answering user queries. 
Interestingly, it provides a higher response quality than feeding the whole document to the LLM on the NaturalQuestions dataset: this could be due to the LLM's ability to generate more relevant answers when given a more focused context, which previous work on context compression has shown~\cite{jiang_longllmlingua_2023}.
Initially low when a first query hits a new document (first question), the \textbf{cache hit rate} increases with repeated queries, amplifying the benefits of caching strategies. 
With a cosine similarity threshold of 0.8 and 200-word contextual summaries, semantic caching achieves a cache hit ratio of 0.2 to 0.3 for datasets with diverse questions, and up to 0.5 to 0.6 for datasets with more similar queries, while maintaining the same quality as answers generated using the full reference document.
Reusing cached  contextual summaries  leads to fewer summaries generated over time, and thus to fewer \textbf{input tokens} ingested by the LLM as the cache grows. In particular, the number of input tokens is significantly lower than when using the full document, at least 50$\%$ lower on average for NaturalQuestions and ArXiv datasets, where questions similarity is high. 
The number of \textbf{output tokens} generated is higher than any other method, but steadily decreases over time, as the cache hit rate increases, and more answers can be generated directly from the cached summaries. 
The higher cache hit rate also leads to a decreasing \textbf{latency} over time, as the system can skip the costly and time-consuming processing of the full document (to answer directly or to generate a new summary). 
Latency breakdown is provided in section~\ref{sec:latency}.

In comparison, \textit{using the full document} consistently achieves highest utility, but incurs high input token usage. Using \textit{non-contextual summaries} starts with higher latency and token usage but quickly stabilizes, though with lower utility.
\textit{Full LLM API caching} appears to provide optimal latency and perfect cache hit rates initially, but what actually happens is that system prompts and retrieved documents overweigh the user query -- the same answer is retrieved regardless of the question asked.
The \textit{No retrieval} method shows varying utility: high for TriviaQA, average for NaturalQuestions, and lowest for ArXiv, indicating the LLM was likely trained on the public datasets but not on recent ArXiv papers.

\subsection{Latency details}
\label{sec:latency}
\begin{figure}[t]
\centering
\includegraphics[width=1.\linewidth, trim=7mm 16mm 3mm 26mm, clip]{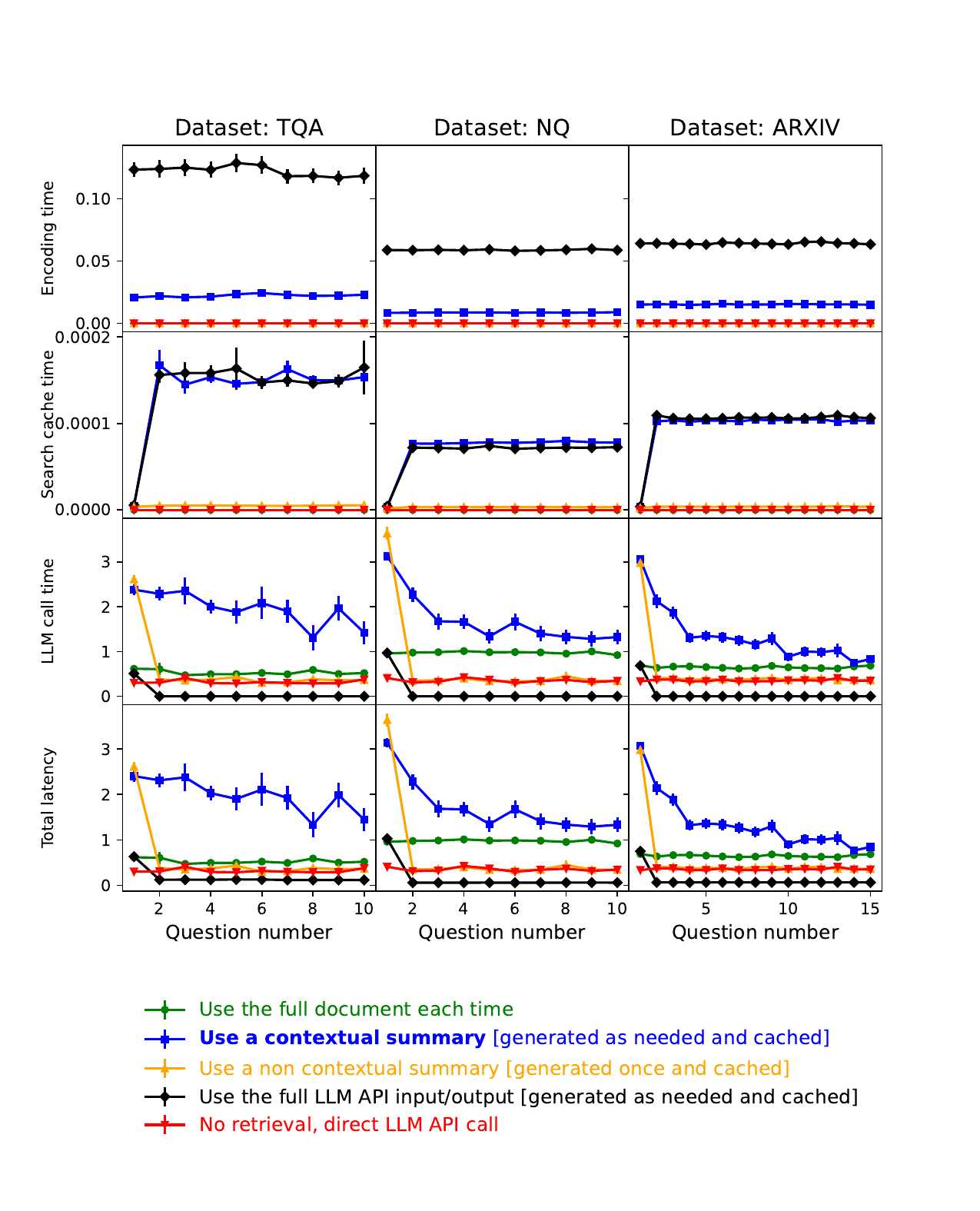}
\vspace{-0.08in}
\caption{LLM API call time, search cache time, encoding time, and total latency of different retrieval methods across the three datasets, for a similarity threshold of 0.8 and summary length of 200 words.}
\label{fig:latency_breakdown}
\end{figure}

Figure~\ref{fig:latency_breakdown} provides a detailed breakdown of the latency of different retrieval methods across the three datasets, for a similarity threshold of 0.8 and summary length of 200 words. The latency is broken down into LLM API call time (the time taken to generate an answer), search cache time (the time taken to search the cache for a relevant summary), and encoding time (the time taken to encode the input tokens). 
The total latency is predominantly ($>99\%$) due to LLM API calls. As the contextual summarization approach may require two API calls (summarization and answering), it can have higher latency than approaches requiring only one call.

\subsection{Varying the similarity threshold and the summary length}
\label{sec:similarity_threshold_summary_length}

\begin{table*}[htbp]
\caption{Performance metrics (cache hit rate, token usage, utility, latency) at different similarity thresholds and summary lengths. Results indicate that the utility depends more the relevance of the reference (higher similarity threshold) than on the length of the summary. The latency, on the other hand, grows with both the sequence length and the similarity threshold.}
\label{tab:stats_output}
\centering
\resizebox{\textwidth}{!}{
\begin{tabular}{l ccc ccc}
\hline
Similarity thresh. & \multicolumn{3}{c}{0.6} & \multicolumn{3}{c}{0.8} \\
Summary length & 100 & 200 & 400 & 100 & 200 & 400 \\
\hline
& \multicolumn{6}{c}{TriviaQA:} \\
\hline
Cache hit rate & 0.61 & 0.61 & 0.61 & 0.27 & 0.27 & 0.27 \\
$\#$ input tokens & 3671 $\pm$ 1116 & 3759 $\pm$ 1115 & 3931 $\pm$ 1110 & 5763 $\pm$ 1380 & 5850 $\pm$ 1378 & 6026 $\pm$ 1371 \\
$\#$ output tokens & 49 $\pm$ 10 & 83 $\pm$ 18 & 149 $\pm$ 34 & 87 $\pm$ 12 & 151 $\pm$ 20 & 277 $\pm$ 37 \\
Utility & 0.51 $\pm$ 0.08 & 0.52 $\pm$ 0.09 & 0.56 $\pm$ 0.08 & 0.70 $\pm$ 0.07 & 0.70 $\pm$ 0.08 & 0.71 $\pm$ 0.07 \\
Total latency & 1.04 $\pm$ 0.15 & 1.31 $\pm$ 0.23 & 1.86 $\pm$ 0.31 & 1.54 $\pm$ 0.19 & 2.12 $\pm$ 0.30 & 3.12 $\pm$ 0.34 \\
\hline
& \multicolumn{6}{c}{NaturalQuestions:} \\
\hline
Cache hit rate & 0.79 & 0.79 & 0.79 & 0.53 & 0.53 & 0.53 \\
$\#$ input tokens & 3557 $\pm$ 298 & 3655 $\pm$ 299 & 3787 $\pm$ 298 & 7461 $\pm$ 1055 & 7556 $\pm$ 1055 & 7723 $\pm$ 1054 \\
$\#$ output tokens & 30 $\pm$ 2 & 51 $\pm$ 5 & 85 $\pm$ 7 & 63 $\pm$ 7 & 108 $\pm$ 12 & 186 $\pm$ 21 \\
Utility & 0.61 $\pm$ 0.04 & 0.64 $\pm$ 0.04 & 0.66 $\pm$ 0.04 & 0.72 $\pm$ 0.03 & 0.74 $\pm$ 0.03 & 0.73 $\pm$ 0.03 \\
Total latency & 0.78 $\pm$ 0.05 & 0.99 $\pm$ 0.06 & 1.32 $\pm$ 0.08 & 1.34 $\pm$ 0.11 & 1.72 $\pm$ 0.16 & 2.44 $\pm$ 0.26 \\
\hline
& \multicolumn{6}{c}{ArXiv:} \\
\hline
Cache hit rate & 0.76 & 0.76 & 0.76 & 0.65 & 0.65 & 0.65 \\
$\#$ input tokens & 1909 $\pm$ 161 & 2024 $\pm$ 161 & 2232 $\pm$ 161 & 2665 $\pm$ 220 & 2780 $\pm$ 220 & 2986 $\pm$ 220 \\
$\#$ output tokens & 39 $\pm$ 4 & 67 $\pm$ 6 & 118 $\pm$ 12 & 53 $\pm$ 4 & 95 $\pm$ 8 & 170 $\pm$ 15 \\
Utility & 0.47 $\pm$ 0.03 & 0.47 $\pm$ 0.03 & 0.49 $\pm$ 0.03 & 0.52 $\pm$ 0.03 & 0.50 $\pm$ 0.03 & 0.51 $\pm$ 0.03 \\
Total latency & 0.81 $\pm$ 0.05 & 1.04 $\pm$ 0.08 & 1.48 $\pm$ 0.12 & 0.99 $\pm$ 0.06 & 1.36 $\pm$ 0.09 & 1.97 $\pm$ 0.14 \\
\hline
\end{tabular}
}
\end{table*}

Table \ref{tab:stats_output} illustrates the impact of varying both the length of contextual summaries and the retrieval similarity threshold.
Increasing the similarity threshold generally enhances answer relevance due to more relevant cached summaries being utilized, albeit with lower cache hit rates. Longer summaries lead to higher utility but also higher input token counts and latency. The summary length does not affect cache hit rate.

\subsection{Picking a similarity threshold: trade-off between utility and cache hit rate}
\label{sec:similarity_threshold_tradeoff}

\begin{figure}[ht]
\centering
\includegraphics{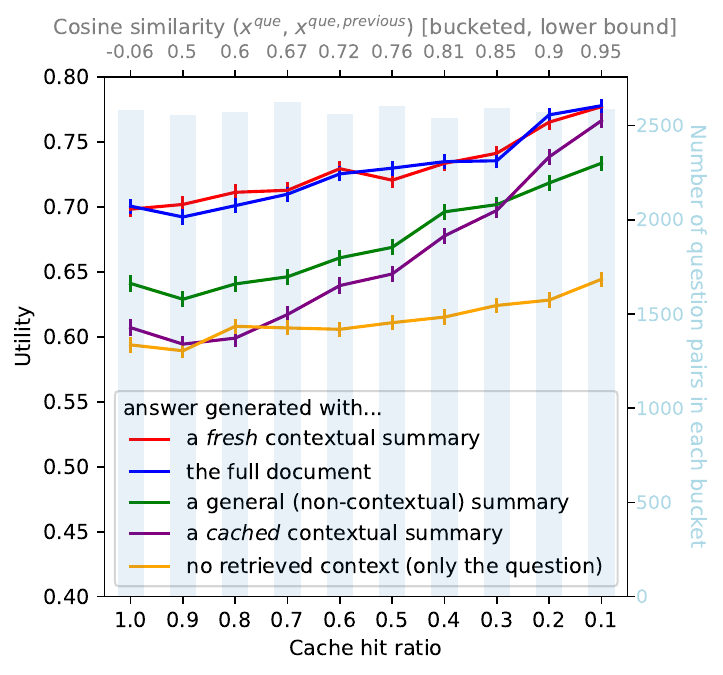}
\vspace{-0.15in}
\caption{NaturalQuestions dataset: End-to-end utility using various versions of the retrieved document as a function of the cache hit ratio. The shaded blue bars (right axis) indicate the number of question pairs per bucket.}
\label{fig:utility_cossim_NQ}
\vspace{-0.1in}
\end{figure}

Selecting an appropriate similarity threshold involves balancing utility with cache hit ratio. Figure~\ref{fig:utility_cossim_NQ} illustrates this trade-off on the NaturalQuestions dataset:

\begin{itemize}
\item Fresh contextual summaries achieve the highest utility, rivaling answers from full documents.
\item Cached contextual summaries perform comparably for high-similarity questions (cosine similarity $>$ 0.85).
\item For low-similarity questions ($<$ 0.6), cached contextual summaries perform worse than general summaries.
\item The distribution of question pairs is skewed toward high-similarity buckets, hinting highly effective caching.
\end{itemize}

For this dataset, a promising threshold lies between 0.85 and 0.9. At 0.9, the system achieves a cache hit ratio of 0.2, improving utility by 4\% over general summaries.

\subsection{Discussion on what influences an answer's generation}
The influence of each component can be measured by checking the agreement (cosine similarity) between the generated answer and the ``ground truth'' answer provided by the dataset. Exploratory experiments showed that, LLM size, temperature, and token length aside, the \textit{question} is the most influential component in answer generation. As question similarity increases, so does answer similarity. The context provided plays a secondary role, while instructions and formatting have the least impact on content (though they affect style and coherence of the answer).

\subsection{Implications for real-world systems}

Semantic caching offers a promising approach to improving cost-efficiency and responsiveness in LLM-based systems, especially in real-world applications such as personal assistants, customer support, and knowledge retrieval. The following trends are expected to continue and amplify over the system's running time, especially for frequently accessed documents: using cached contextual summaries achieves \textbf{high utility} comparable to full-document answers while benefiting from an \textbf{improved cache hit rate} as repeated queries on similar topics occur over time. As more summaries are reused, the number of input tokens decreases, \textbf{lowering resource consumption and operational costs}. By avoiding redundant computations, \textbf{response times are minimized}. These trends suggest that semantic caching of contextual summaries is a viable and effective strategy for LLM-based systems, particularly when handling recurring queries or frequently accessed documents.

\subsection{Challenges and limitations}

Despite its potential, semantic caching faces several challenges: new documents or dissimilar queries require fresh computations (\textbf{cold-start problem}); determining the \textbf{optimal similarity threshold} that balances utility and cache hit rate remains domain-specific and requires careful tuning; as the system \textbf{scales} to handle larger datasets, growing demand, and more complex queries, maintaining efficient cache management becomes critical; cached summaries may contain sensitive information, making it difficult to share caches among users securely and raising important \textbf{privacy concerns} that must be addressed in multi-user environments.

\section{Conclusion and future work}

In this study, we demonstrate that caching intermediate contextual summaries can significantly reduce computational costs of LLM-based QA systems while maintaining accuracy. By balancing efficiency and utility through selective caching, our approach lays the groundwork for cost-effective and scalable LLM applications, particularly in environments where mobile devices and cloud servers must work together.

To unlock its full potential, future research should address:

\paragraph{Technical enhancements} Developing adaptive algorithms to optimize similarity thresholds based on query patterns and exploring advanced summarization techniques to optimize information preservation.

\paragraph{Scalability and real-world deployment} Handling larger datasets and multi-user systems, and tailoring techniques for specific industries to maximize relevance.

\paragraph{Privacy concerns} Implementing mechanisms such as differential privacy, user-specific caches, or encrypted summaries to address privacy in multi-user environments.

\paragraph{Broader applications} Accessing corporate datasets to model realistic query distributions and extending the system to handle queries requiring synthesis of multiple documents.

Selective semantic caching of intermediate outputs has the potential to transform LLM-based systems by enabling cost-efficient, scalable, and adaptive solutions for a wide range of applications. By addressing these challenges, semantic caching can evolve into a robust framework offering significant benefits for real-world deployments.

\bibliographystyle{plain}
\bibliography{main}

\end{document}